%% file: root.tex
\documentclass[letterpaper, 10 pt, conference]{ieeeconf}  

\IEEEoverridecommandlockouts                              

\usepackage{graphicx}
\usepackage{amsmath} 
\usepackage{amssymb}  
\usepackage{xcolor}
\usepackage{multicol}
\usepackage{array}
\usepackage{booktabs}
\usepackage{cite}
\usepackage{float}
\usepackage{url}
\usepackage{hyperref}
\usepackage[load-configurations=version-1]{siunitx}
\usepackage{subcaption} 

\input{math_command.tex}

\newcommand{\imgcell}[1]{\includegraphics[scale=0.5]{#1}}
\title{\LARGE \bf
 Action-conditioned Deep Visual Prediction with RoAM, a new Indoor Human Motion Dataset for Autonomous Robots
}

\author{Meenakshi Sarkar$^{1}$, Vinayak Honkote$^{2}$, Dibyendu Das$^{2}$ , Debasish Ghose$^{1}$
\thanks{$^{1}$ Aerospace Engineering Dept,
        Indian Institute of Science, Bangalore 560012
        {\tt\small \{meenakshisar, dghose\}@iisc.ac.in}}%
\thanks{$^{2}$Intel India
        {\tt\small \{vinayak.honkote, dibyendu.das\}@intel.com}}%
}

\begin{document}

\maketitle
\thispagestyle{empty}
\pagestyle{empty}

\begin{abstract}
With the increasing adoption of robots across industries, it is crucial to focus on developing advanced algorithms that enable robots to anticipate, comprehend, and plan their actions effectively in collaboration with humans. We introduce the Robot Autonomous Motion (RoAM) video dataset, which is collected with a custom-made turtlebot3 Burger robot in a variety of indoor environments recording various human motions from the robot's ego-vision. The dataset also includes synchronized records of the LiDAR scan and all control actions taken by the robot as it navigates around static and moving human agents. The unique dataset provides an opportunity to develop and benchmark new visual prediction frameworks that can predict future image frames based on the action taken by the recording agent in partially observable scenarios or cases where the imaging sensor is mounted on a moving platform. We have benchmarked the dataset on our novel deep visual prediction framework called ACPNet where the approximated future image frames are also conditioned on action taken by the robot and demonstrated its potential for incorporating robot dynamics into the video prediction paradigm for mobile robotics and autonomous navigation research.

\end{abstract}

\section{Introduction}
Recently the world has witnessed the rise of data-driven models with the development of large-scale generative AI systems. Despite their demonstrated ability to generalize over a large spectrum of scenarios, research has repeatedly emphasized the significance of accurately annotated and balanced datasets, as they play a crucial role in determining the performance and efficacy of large-scale models (\cite{dataset_imbalance1_2020}, \cite{dataset_balance2_2017}). While self-attention-based  transformer models \cite{Vaswani_attention2017} have set a new standard for language prediction and generation, the same cannot be said for video and image prediction models. This is because video data is much larger in dimension and has more stochasticity, making it challenging to model accurately. 
\begin{figure}[t]
\centering
  \includegraphics[width=0.9\linewidth]{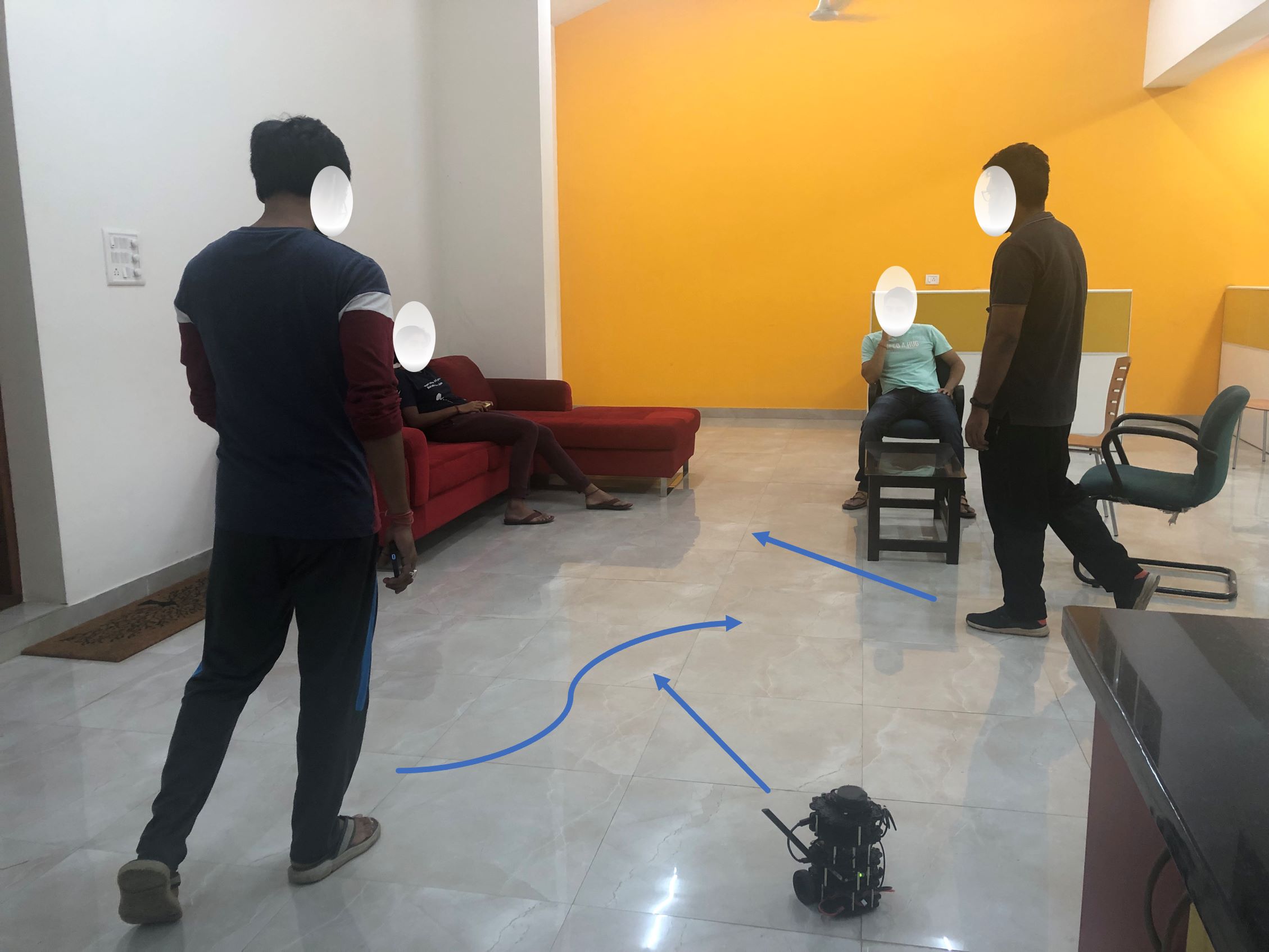}
  \caption{Prediction is the first step towards intelligent planning}
\label{fig:roman_intro1}
\end{figure}

This problem becomes even more complicated when the camera is mounted on a mobile platform as in the case of autonomous mobile robots \cite{villegasNeurIPS2019}. Here the prediction algorithm will have to also model the changing background sceneries as well as the multiple moving objects in the foreground. The computer vision community also refers to this problem as the partially observable visual prediction problem where the dynamics of the mobile platforms are entangled with the predicted future image frames. Despite the challenges in modelling video data accurately, it is imperative for  the robotics community to solve this problem. This is because prediction is a necessary step in planning, and in order to design an optimal motion planner, a powerful state prediction mechanism is needed.

The scenario in Figure \ref{fig:roman_intro1} shows a robot moving in an indoor space having multiple moving and static human agents in the same environment, whose probable trajectories are also shown. The robot needs to plan its path and navigate through this environment while avoiding collisions with humans. Here, a reliable visual prediction framework could greatly assist the robot's planning algorithm. However, due to the challenges and uncertainties in such environments, visual prediction in partially observable scenarios, although addressed to some extent in the existing literature, still remains an open problem.

In these scenarios, one of the main challenges in visual prediction arises due to the interdependence between the robot's movements and the visual data captured by its camera. Each pixel of data recorded by the camera sensor is conditioned on the dynamics of the robot. In the past, researchers have attempted to tackle this problem in a variety of ways. However, with the availability of high compute power, there is a recent trend in generating high fidelity video predictions with various generative architectures such as Generative Adversarial Networks (GAN) and Variational Autoencoders (VAE) \cite{LiangICCV2017, Denton, BabaeizadehICLR2018, LeeICLR2018, CastrejonICCV2019, GaoCVPR2020, villegasNeurIPS2019, slrvp2020}. Recently there has been a lot of interest in designing efficient visual transformers for video prediction tasks \cite{video_transformer2022}. for  Several recent studies, including those by Gao et al. \cite{GaoCVPR2020} and Villegas et al. \cite{villegasNeurIPS2019}, have attempted to tackle the issue of partial observability in dynamic scene forecasting through the use of generative models such as GAN and VAE. Despite their ability to produce realistic predictions for moving camera scenarios, these approaches demand significant computational resources, rendering them impractical for deployment on low- to mid-tier robotic platforms. Sarkar et al. \cite{sarkar2021} proposed  Velocity Acceleration Network (VANet), to extract and disentangle the dynamics of the robot's motion from the visual data recorded by its camera. Their approach utilized motion flow encoders to estimate the relative motion kernels between the camera's reference frame and the environment, effectively approximating the first and second-order dynamics. However, none of the above approaches for predicting image frames from moving cameras explicitly model the dynamics of the mobile platform. Instead, they infer the effect of the robot's motion from the raw data itself using learned priors in the case of stochastic architectures like VAEs \cite{Denton}, \cite{LeeICLR2018}, or through kernels from the motion encoders in the case of deterministic networks like VANet \cite{sarkar2021} and MCNet \cite{villegas}. 

In this paper, we work on the premise that a more sensible approach would be to incorporate the control action data from the recording platform directly into the video prediction framework and condition the predicted image frames on the actions taken by the mobile agent for better accuracy. Including control action data in the video prediction framework is consistent with standard practices in control systems engineering for autonomous navigation since such data is often available. However, existing video datasets like KITTI \cite{KITTI} or KITTI-360 \cite{KITTI-360}, A2D2 \cite{A2D2} do not offer access to the synchronised control action or driving data of the car on which the camera was mounted during recording.

With the above as a motivation, we introduce Robot Autonomous Motion Dataset (RoAM) a novel open-source stereo-image dataset captured using a Zed mini stereo camera mounted on a Turtlebot3 Burger robot. RoAM includes timestamped and synchronized control action data recorded while the robot navigated indoor environments. The dataset mainly consists of human motion data, as the robot explores different environments autonomously and avoids collisions with multiple human agents in its surroundings. RoAM also contains timestamped data from the 2D LiDAR scan of an LDS-01 sensor and the IMU and odometry data from the robot.

Furthermore, we present ACPNet, a new action-conditioned video prediction framework that conditions predicted image frames on robot action. Finally, we provide the benchmarking results of ACPNet on the new RoAM dataset and compare its performance with its existing counterparts in the literature. In the following section, we briefly discuss the existing datasets for deep visual prediction frameworks and compare the novelty of RoAM.

The paper is structured into the following sections: Section \ref{sec:literature_dataset} presents an in-depth literature review of currently available datasets for video prediction tasks and emphasizes the necessity of a new action-conditioned dataset. Subsequently, Section \ref{sec:data_description} provides a comprehensive overview of the data collection process for the proposed RoAM dataset, including hardware specifications and processed data. The mathematical framework of the proposed ACPNet is discussed in detail in Section \ref{sec:ACPNet}, followed by an analysis and discussion of the ACPNet's performance on the RoAM dataset in Section \ref{sec:results}. Finally, Section \ref{sec:conclusion} concludes the paper with a few remarks regarding the possible future research directions that can be pursued using this newly developed dataset.

\section{Literature review on similar dataset papers and why RoAM is novel} \label{sec:literature_dataset}
In the literature on video prediction frameworks, Srivastava et al. \cite{srivastava} made a seminal contribution to visual prediction through the use of deep recurrent neural networks. Their work involved introducing the dancing MNIST dataset and benchmarking their network on human action recognition and classification tasks on the UCF-101 \cite{UCF-101} and HMDB-51 \cite{HMDB-51} datasets. However, these datasets are stationary in nature, recorded from a stationary camera, and mostly consist of videos available on the internet. With the success of Convolutional LSTM architectures \cite{xing}, there has been a surge of research interest in designing better and more accurate video prediction networks. This has led to the development of new, more detailed, and complicated video datasets to test and analyze these frameworks. The KTH human action dataset \cite{kth} introduced the complexities of the structured and linked motion of six different types of human actions in grey and static backgrounds. The BAIR Push dataset \cite{finn2, LeeICLR2018} introduced the stochastic elements that arise from the interactions between a manipulator arm and various soft and rigid objects within its workspace. However, neither of these datasets offered a partially observable scenario, as in both cases, the camera was stationary. 

\begin{table}[t]
\centering
\caption{List of video Datasets}
\begin{tabular}[t]{>{\raggedright}p{0.174\linewidth}>{\raggedright}p{0.12\linewidth}>{\raggedright\arraybackslash}p{0.28\linewidth}>{\raggedright\arraybackslash}p{0.23\linewidth}}
\toprule
&Mode of Capture &Data Collected& Application\\
\hline
Dancing MNIST \cite{srivastava}& Fixed Camera&Generated Images& Video Prediction and generation.\\
\hline
KTH \cite{kth}&Fixed Camera&Monocular and Grayscale& Human Action detection and video prediction.\\
\hline
UCF-101 \cite{UCF-101}, HMDB-51 \cite{HMDB-51}&Fixed Camera&Monocular and color images & Human Action detection and video prediction.\\
\hline
BAIR Push \cite{finn2, LeeICLR2018}&Fixed Camera&Monocular and colour images & Manipulator action conditioned video data for object manipulation task.\\
\hline
KITTI \cite{KITTI}, KITTI-360 \cite{ KITTI-360}& Moving Camera&Stereo and 360-degree panoramic colour images and 3D point cloud, \textbf{No action data} & autonomous driving and robotics.\\
\hline
Caltech Pedestrian Dataset \cite{CalTech-pedestrian}& Moving Camera&Stereo colour images, \textbf{No action data} & autonomous driving and robotics.\\
\hline
A2D2 \cite{A2D2}& Moving Camera&Stereo color images with 3D point cloud, \textbf{No action data}& autonomous driving and robotics, \textbf{No action data}.\\
\hline
Stanford Go \cite{Stanford-Go}& Moving Camera&Stereo color fisheye images with 2D laser scan and pose, \textbf{No action data} & autonomous indoor robotics.\\
\hline
\textbf{RoAM}& Moving Camera&Stereo color images with 2D LiDAR scan, Odometry and \textbf{robot action data} & autonomous indoor robotics.\\
\hline
\bottomrule
\end{tabular}\label{tab:1}\vspace{-1em}
\end{table}%
The emphasis on partially observable scenarios emerged alongside the progress of research in autonomous driving. Datasets like KITTI \cite{KITTI} and CalTech Pedestrian \cite{CalTech-pedestrian} are from video footage captured from a vehicle driving through regular traffic in urban environments. Recently Audi introduced their A2D2 or Audi Autonomous Driving Dataset \cite{A2D2} which comprises 43,424 stereo image pairs with corresponding high-resolution 3D point clouds, as well as 6-second-long HD video sequences captured from six cameras mounted on vehicle driving in various urban and rural environments. KITTI dataset is also extended to include panoramic 360-degree views and surround-view perception with the  new KITTI-360 \cite{KITTI-360} dataset. All the datasets mentioned are widely used in autonomous driving research for various tasks such as perception, prediction, and planning, as they are heavily annotated for object detection and segmentation benchmarks. However, none of them pertains to indoor robot navigation and motion planning tasks. The Go Stanford dataset \cite{Stanford-Go} offers stereo image pairs from a fisheye camera, 2D laser scans, odometry, and pose estimation data collected from an indoor robot, which can be used for various mobile robotics applications. However, the dataset's low frame rate of only 3fps makes it unsuitable for video prediction frameworks. Additionally, none of the above-mentioned datasets provides the action or actuator control  data so that video prediction frameworks can be trained and conditioned on them.

Similar to the Go Stanford data, RoAM is also an indoor navigation dataset consisting of various long corridors and lobby spaces inside. RoAM provides timestamped stereo image pairs from a Zed mini camera recorded at a frame rate of 15 fps and the timestamped and synchronised control action data derived from the Turtlebot's autonomous navigation module. The accuracy of the timestamping and synchronization of the multi-modal dataset further enhances its utility for precise analysis and testing of various visual prediction frameworks pertaining to autonomous navigation. A detailed description of the dataset along with a description of the data collection process and the platform is given in the following section. We have also provided the utility and recording modes of different video datasets in Table \ref{tab:1} 

\section{Data collection setup, hardware description and Data Structure}\label{sec:data_description}
\begin{figure}[h]
\centering
  \includegraphics[width=0.85\linewidth]{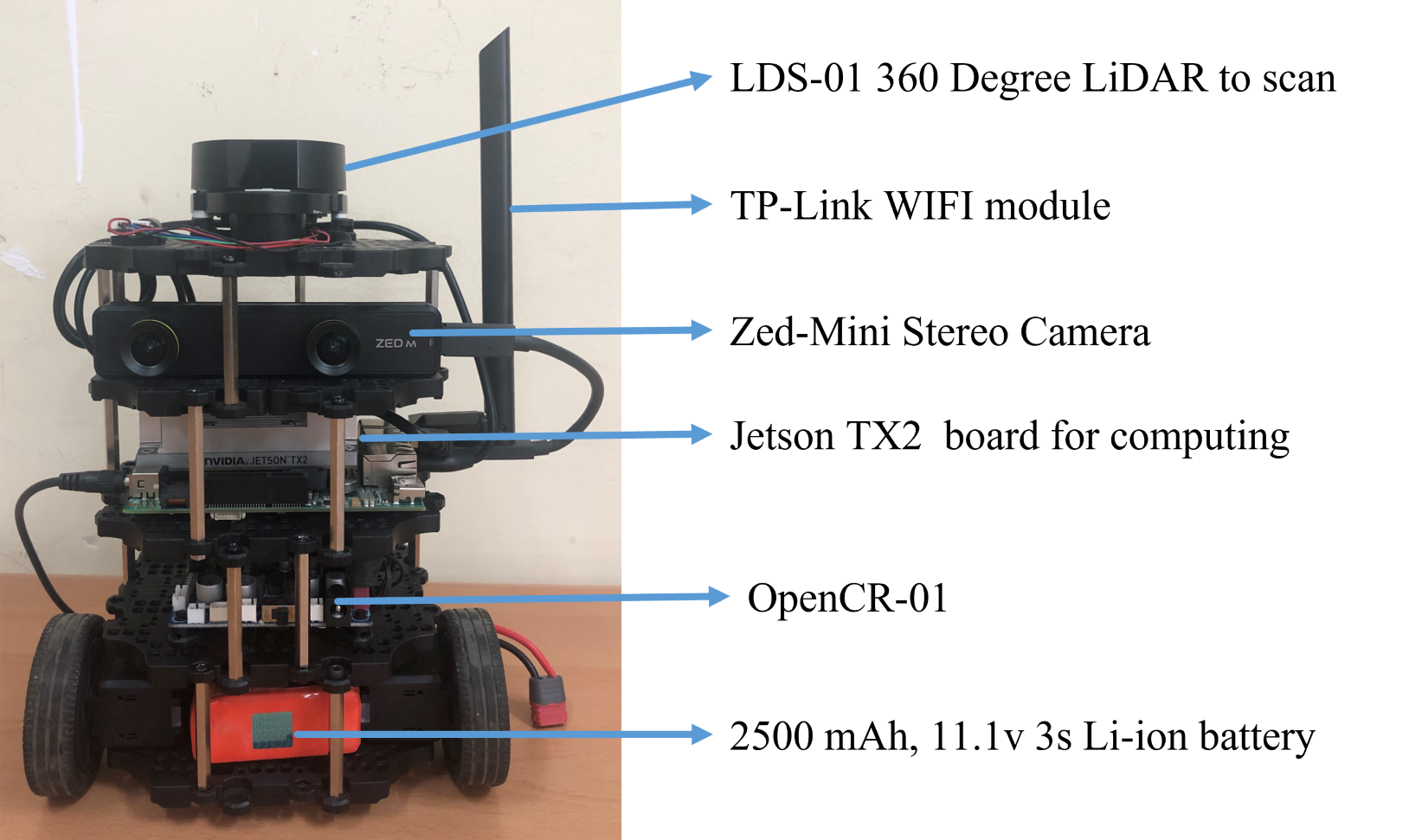}
  \caption{Burger hardware with its various components and sensors. This custom made Turtlebot3 robot is used for the collection of RoAM.}
\label{fig:roman_burger}
\end{figure}
The RoAM dataset is collected using a custom-built Turtlebot3 Burger robot. We have used Zed mini stereo vision camera for capturing the left and right timestamped image pairs. Other than that the robot is equipped with an LDS-01 2-dimensional LiDAR, a TP-link WiFi communication module as shown in Figure \ref{fig:roman_burger}. The Turtlebot3 employs two DYNAMIXEL XL430-W250 servo motors for navigation, utilizing current-based torque control. These motors are actuated and controlled by the OpenCR-01 board, which is integrated into the platform. For our specific application, we have selected the Jetson TX2 board as the on-board computer, operating on the ROS Melodic framework \cite{ROS} and the Ubuntu 18.04 operating system. This setup offers the advantage of leveraging the Jetson TX2's high computational power to support complex robotic tasks, such as perception, navigation, and machine learning. 
\begin{figure}[h]
\centering
  \includegraphics[width=0.95\linewidth]{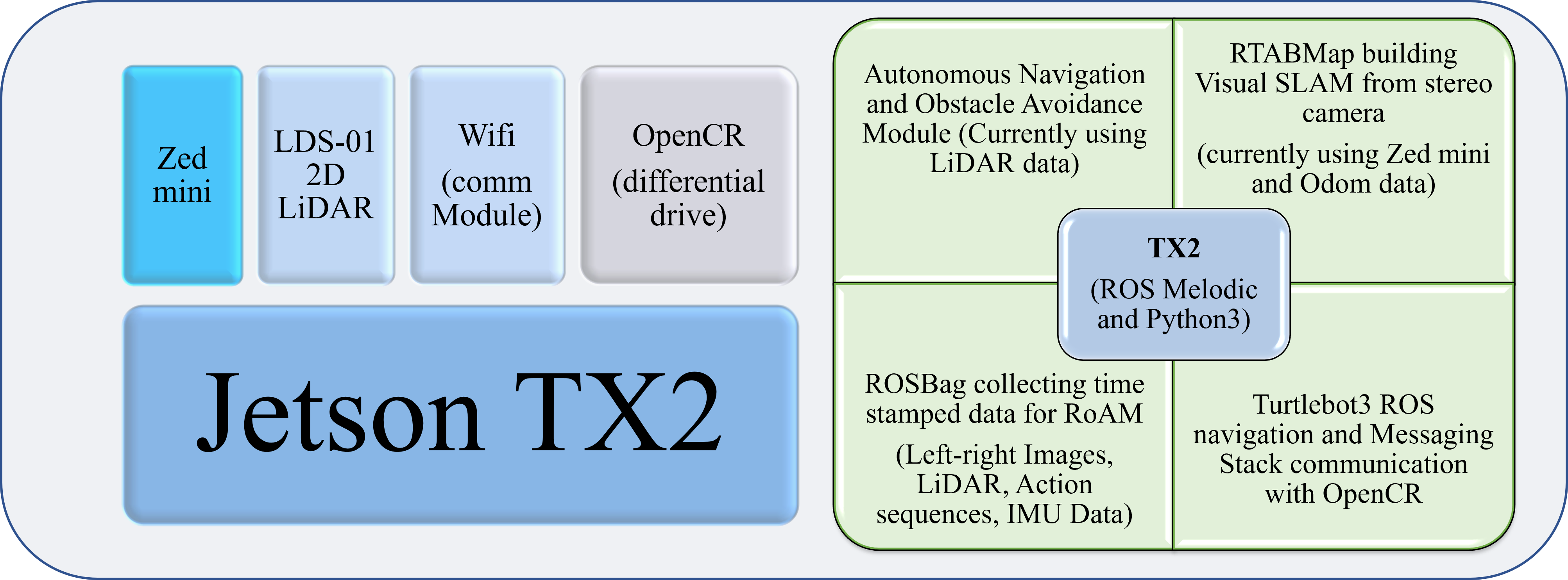}
  \caption{RoAM software-hardware stack. Here we have shown the integration between the various sensors and hardware components with the Jetson TX2 board on one side and on the other, we show the various software modules that are inter-connected via ROS.}
\label{fig:roman_RoAM_stack}
\end{figure}

Figure \ref{fig:roman_RoAM_stack} provides a visual representation of the software stack for navigation and data collection in RoAM. The TX2 board serves as the central computing unit, integrating both LiDAR and camera sensors. Although the robot can operate and explore its surroundings autonomously, a WiFi communication link is maintained with the ground station for safety purposes only. The entire software stack for Burger in Figure \ref{fig:roman_RoAM_stack} can be broadly categorized into four main functionalities: (i) Autonomous navigation and obstacle avoidance stack, (ii) Data collection using ROSBag and Zed wrapper, (iii) Visual SLAM Module, and (iv) the core Turtlebot3 ROS navigation and messaging stack for the OpenCR board. 
\begin{figure}[h]
\centering
  \includegraphics[width=0.85\linewidth]{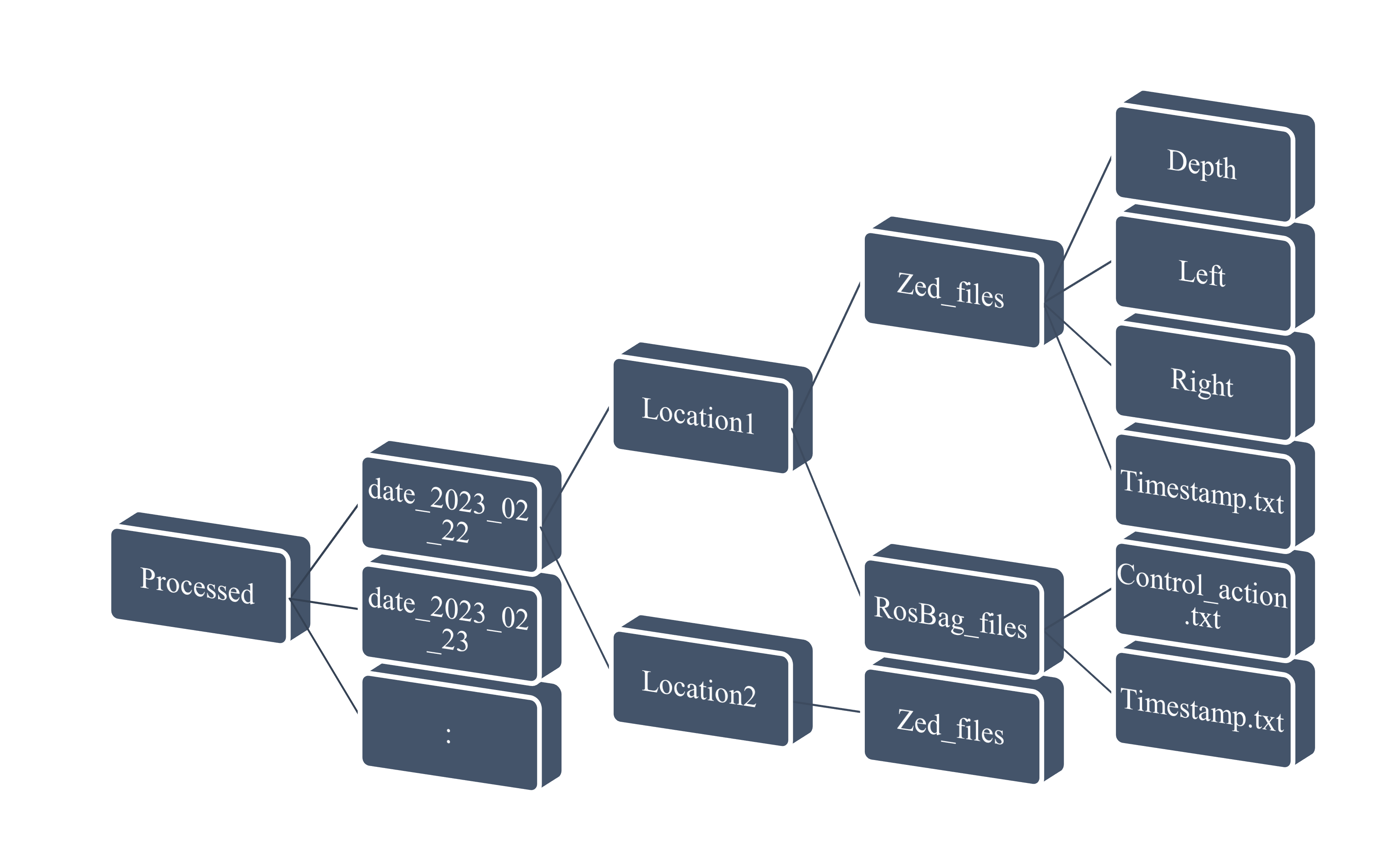}
  \caption{The processed data file structure of RoAM}
\label{fig:roman_RoAM_processed_data}
\end{figure}

\begin{figure*}[t]
    \centering
    \includegraphics[width=0.8\textwidth]{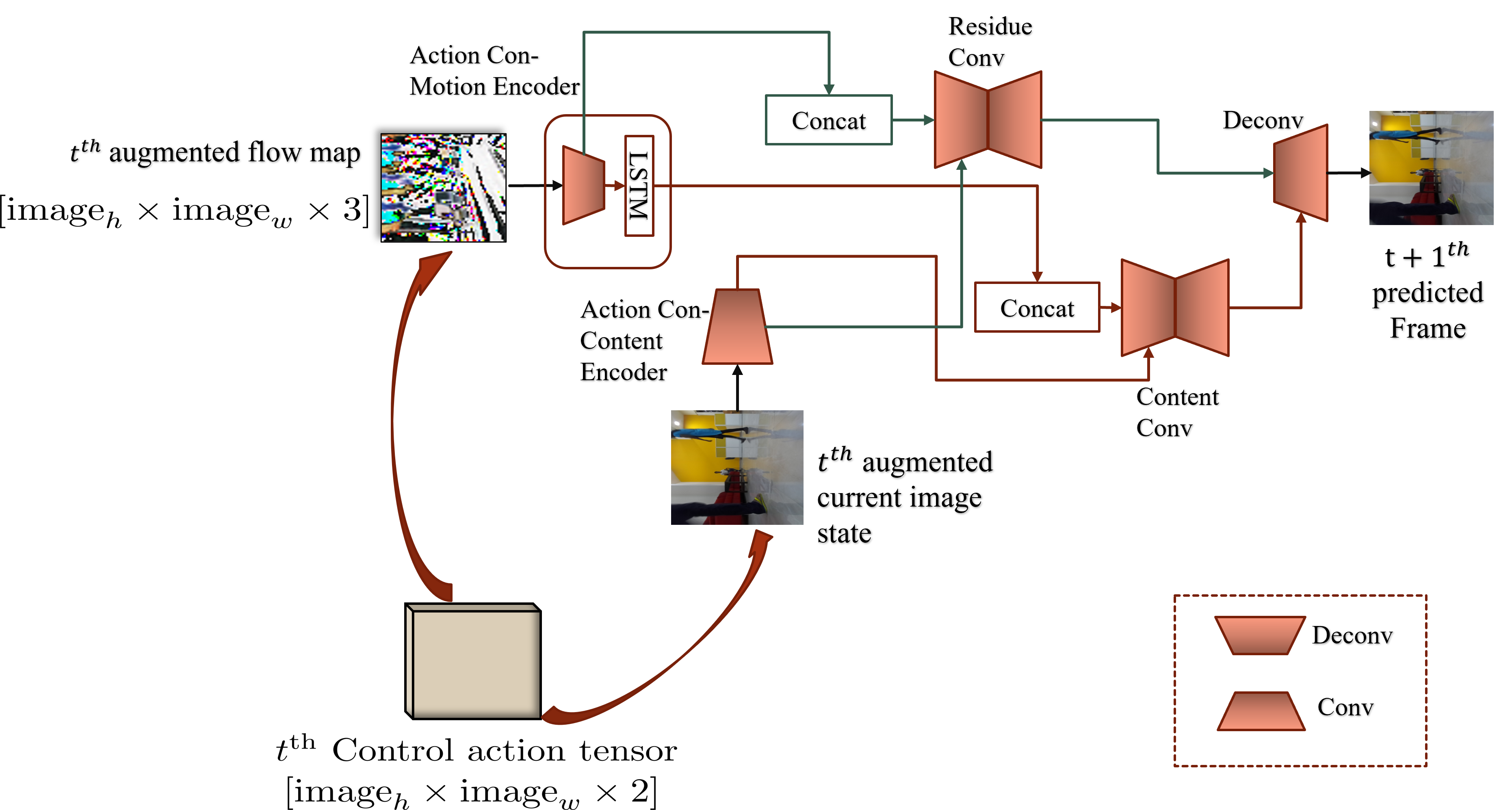}
    \caption{Architecture of ACPNet consists of an augmented motion and content encoder. The network uses augmented current image state and motion flow maps to learn the complex spatio-temporal interdependencies between the robot dynamics and the observed image pixels. It also uses a residue convolution and a decoder module to generate the predicted image frames.}
    \label{fig:ACPNet}
\end{figure*}
For the exploration and data collection phase, the autonomous navigation and collision avoidance module is designed utilizing the 2D scans provided by the LiDAR sensor. Although there is considerable literature available on collision avoidance algorithms utilizing LiDAR data \cite{sarkar-cis2023, michels_2005}, we have designed and executed a Collision Cone-based approach \cite{collision_cone} for obstacle avoidance. To gather data for the RoAM dataset, we have allowed the robot to freely navigate through various areas of the building while recording LiDAR scans, IMU, Odometry, and timestamped Control action data in the ROSBag format. Additionally, stereo image pairs are captured using the Zed mini camera and saved in the proprietary \textit{.svo} format provided by Stereolabs, which are later processed and saved in \textit{.png} format. While the robot also has the capability to generate 3D point cloud data through the structure from motion (SfM) or visual SLAM module, we have kept this feature as a potential future extension of the RoAM dataset.

Figure \ref{fig:roman_RoAM_processed_data} provides a visual depiction of the post-processed dataset, which includes timestamped and synchronized data. The dataset was collected by recording approximately 38,000 stereo image pairs over a period of 5 days in 8 different corridors and indoor lobby spaces comprising 25 different recorded sequences located within a building with corridors, lobby spaces, staircases, offices and laboratories, that experience frequent human movement. The total experimental time was approximately 6 hours. In addition to the left and right time-stamped image pairs, we also provide the depth or disparity map, which was obtained using the Zed mini's camera calibration data, in a 32-bit single-channel \textit{.tiff} format, as illustrated in Figure \ref{fig:roman_RoAM_processed_data}. The dataset primarily contains a variety of human movements, including walking, jogging, hand-waving, gestures, and sitting actions, which any small robot similar to Turtlebot3 may encounter while exploring indoor environments populated by humans. The image frames in the RoAM dataset also contain various artifacts that are commonly observed in real-world scenarios, such as reflections, glare, and motion blur.   
 
When it comes to the control action data, Turtlebot3 is a non-holonomic robot and has only two actuator actions for controlling 3 degrees of freedom: ($x, y$ and yaw). The available two control actions are the forward velocity along the $x$-coordinate of the robot's body frame and the rotational rate or turn rate about the body $z$- axis. For our experimental purposes, we have kept the forward velocity within the range of $[0,0.1]$ m/s and the rotational rate within $[-1.8, 1.8]$ rad/sec. For every recording session, the robot starts from a stationary state with zero forward velocity and gradually accelerates to the allowed maximum velocity  in case there are not obstacle present in the front. The maximum allowed forward acceleration is set to $0.2$ m/s$^2$. The final timestamped control action data are saved in a corresponding \textit{.txt} file for each recorded sequence following the file structure given in Figure \ref{fig:roman_RoAM_processed_data}. In the following section, we will describe our prediction framework, ACPNet, which utilizes the action data to condition each predicted pixel, creating a dependent and coupled relationship between the two.
\section{Action Conditioned Prediction Network}\label{sec:ACPNet}

This paper presents the Action-Conditioned Prediction framework, ACPNet, which draws inspiration from the VANet architecture proposed by Sarkar et al. \cite{sarkar2021}. We have modified the VANet architecture to incorporate the last action taken by the robot as a condition for predicting future image frames. While Sarkar et al. \cite{sarkar2021} presented a comprehensive mathematical framework for their physics-based second-order network, the details of how robot actions could be integrated into the same or a similar architecture was not provided. We have developed a novel method for addressing this issue with ACPNet. Our approach utilizes an augmented current image state and augmented pixel difference maps to establish the relationships and dependencies between predicted image frames and robot control actions. This method incurs no significant computational burden on architectures such as VANet \cite{sarkar2021} or MCNet \cite{villegas}, as we demonstrate below through a  mathematical framework and analysis. A pictorial representation of the network is given in Figure \ref{fig:ACPNet} 

Assume that $x_t$ is the current image frame of dimension $[\image_h\times\image_w\times 3]$ and $o_t$ is the current first-order flow map of dimension $[\image_h\times\image_w\times 3]$ where
\begin{equation}
    o_t=x_t-x_{t-1}
    \label{eq:flow_map}
\end{equation}
Additionally, we have access to the previous $t$ image frames represented as $[x_{t:1}]$, $t-1$ first-order flow maps represented as $[o_{t-1:2}]$, where $o_i=x_i-x_{i-1}$ and a list of control actions at $t$ timesteps represented as $a_{t:1}$, $a_t \in \sR^{m}$, where $m$ is the number of available of control actions. For Turtlebot3, $m=2$. Our network, ACPNet, is parameterized by $\Theta$ and Our objective of the find the optimal $\Theta^*$ such that:
\begin{equation}
    \Theta^*=\underset{\Theta}{\text{argmin}} |x_{t+1}-\tilde{x}_{t+1}|^{L_p}
    \label{eq:cost_optimization}
\end{equation}
where $\tilde{x}_{t+1}$ is the predicted future frame by the ACPNet and can be written as:
\begin{equation}
    \tilde{x}_{t+1}=\sF_{ACPNet}(x_t,o_t,a_t|\Theta, x_{t-1:1},o_{t-1:2},a_{t-1:1})
    \label{eq:acpnet}
\end{equation}
In contrast to VANet where Sarkar et al. \cite{sarkar2021} employed two distinct motion encoding modules, namely the Velocity and Acceleration Encoders, to encode motion information from first and second-order pixel difference maps, respectively, ACPNet employs a single motion encoding module that takes augmented first-order flow maps $\hat{o}_t$ as inputs. Similar to the approach in Sarkar et al. \cite{sarkar2021} and Villegas et al. \cite{villegas}, we use convolutional LSTM modules \cite{convlstm} in our motion encoder to capture and model the temporal dependencies within consecutive image frames. For analytical purposes, we divide the parametric space $\Theta$ into two components, $\Theta=[\Theta_R, \Theta_F]$, where $\Theta_R$ represents the parametric space associated with the recurrent architecture of the Motion Encoder, and $\Theta_F$ comprises the rest of the parametric space of the feed-forward architecture.

The augmented current image state $\hat{x}_t$ and flow map $\hat{o}_t$ are generated as follows:
  \begin{equation}
   \hat{x}_t=[x_t, \alpha_t]  
  \label{eq:x_hat}
\end{equation}
\begin{equation}
    \hat{o}_t=[o_t, \alpha_t]
    \label{eq:o_hat}
\end{equation}
respectively, where $\alpha_t$ is a normalized action map of dimension $[\image_h\times\image_w\times m]$. This is to establish the dependency between each observed pixel from the camera and the actions taken by the robot at time $t$. 

The motion encoding kernel $\tilde{f}_t$ generated by the motion encoder module at timestep $t$ is conditioned on the past flow maps $(o_{t:2})$ and robot control actions up to time $t$ $(a_{t:1})$, and can be represented as follows:
\begin{equation}
    \tilde{f}_t=\sF_{MoEnc}(o_t,a_t|\Theta_R, o_{t-1:2},a_{t-1,2})
    \label{eq:MoEnc}
\end{equation}
Using  (\ref{eq:o_hat}) we can rewrite  (\ref{eq:MoEnc}) as :
\begin{equation}
    \tilde{f}_t=\sF_{MoEnc}(\hat{o}_t|\Theta_R, \hat{o}_{t-1:2})
    \label{eq:MoEnc_mod}
\end{equation}
where, $\hat{o}_i=[o_i, \alpha_i]$ $\forall i\in[2,t-1]$. In addition, we shall represent the initial state of our prediction problem by $\Omega_0=\hat{o}_{t-1:2}$, as these augmented flow maps from the previous $t-1$ time steps are utilized for initializing the motion encoding module. Thus,  (\ref{eq:MoEnc_mod}) can we further modified as :
\begin{equation}
    \tilde{f}_t=\sF_{MoEnc}(\hat{o}_t|\Theta_R, \Omega_0)
    \label{eq:MoEnc_mod_omega}
\end{equation}
Once $\tilde{f}_t$ is generated, the predicted future image frame at time step $t+1$ can be generated with the augmented current image state $\hat{x}_t$ as:
\begin{equation}
    \tilde{x}_{t+1}=\sF_{\overline{MoEnc}}(\hat{x}_t, \tilde{f}_t|\Theta_F, \Omega_0)
    \label{eq:MoEnc_bar}
\end{equation}
where, $\sF_{\overline{MoEnc}}$ denotes the rest of the feed-forward architectures parameterised by $\Theta_F$. Without the loss of generality,  (\ref{eq:MoEnc_mod_omega}) and  (\ref{eq:MoEnc_bar}) can be combined and written as:
\begin{equation}
    \tilde{x}_{t+1}=\sF_{MoEnc+\overline{MoEnc}}(\hat{x}_t, \hat{o}_t|\Theta_F,\Theta_R, \Omega_0)
    \label{eq:ACPNet_mod}
\end{equation}
using  (\ref{eq:x_hat}) and  (\ref{eq:o_hat}) we can rewrite  (\ref{eq:ACPNet_mod}) as:
\begin{equation}
    \tilde{x}_{t+1}=\sF_{ACPNet}(x_t, {o}_t, \alpha_t|\Theta, \Omega_0)
    \label{eq:ACPNet_mod2}
\end{equation}
Here one can easily see that  (\ref{eq:ACPNet_mod2}) is a slightly generalised version of  (\ref{eq:acpnet}). Therefore, by utilizing Eqns. (\ref{eq:MoEnc})-(\ref{eq:ACPNet_mod2}), we have demonstrated how our approach of augmenting image states and flow maps with control action maps can effectively incorporate the actions executed by the robot to predict future frames, with minimal modifications to the original architecture. Apart from the motion encoding component, we have made minimal alterations to the content, convolution, residual, and decoder modules of VANet to integrate them into the feed-forward element of ACPNet. We only made minor adjustments to align the modified input dimensions with the content encoder. Figure \ref{fig:ACPNet} illustrates how the action map was augmented with the current image and motion flow maps during the prediction or inference stage of the network. 

\begin{figure*}[ht]
\centering
\subcaptionbox{\label{fig:roam_quant_vgg16}}{\includegraphics[width=0.23\textwidth]{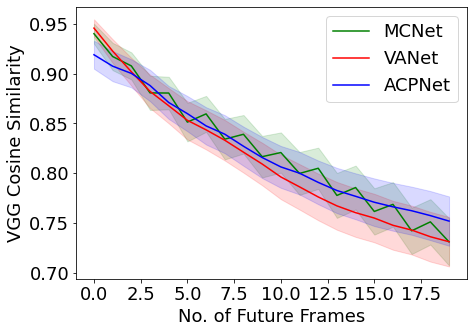}}%
\hfill 
\subcaptionbox{\label{fig:roam_quant_fvd}}{\includegraphics[width=0.23\textwidth]{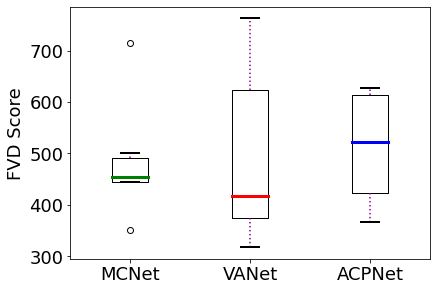}}%
\hfill 
\subcaptionbox{\label{fig:roam_quant_ssim}}{\includegraphics[width=0.23\textwidth]{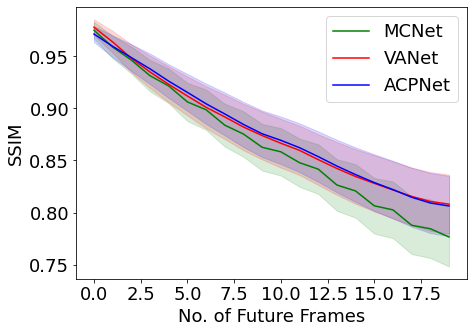}}%
\hfill 
\subcaptionbox{\label{fig:roam_quant_psnr}}{\includegraphics[width=0.23\textwidth]{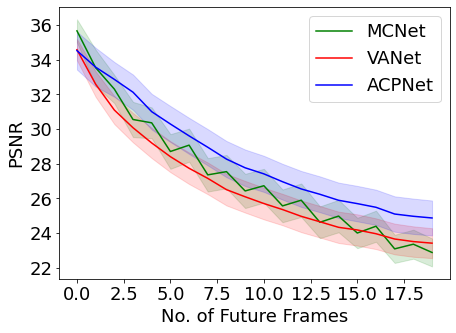}}%
\caption{A Frame-wise quantitative analysis of ACPNet, VANet and MCNet on RoAM dataset for predicting 20 frames into the future based on the past history of 5 frames. Starting from the left, we have plotted the mean performance index for VGG 16 Cosine Similarity (\textbf{higher is better}), FVD score (\textbf{lower is better}), SSIM (\textbf{higher is better}) and PSNR (\textbf{higher is better}) on the test set. }
\label{fig:roam_quant}
\end{figure*}

The ACPNet is trained using reconstruction and gradient difference losses as described in Sarkar et al. \cite{sarkar2021}. Our multi-timestep prediction inference process followed an iterative sequential approach, as also discussed in Sarkar et al. \cite{sarkar2021}. Specifically, in approximating multiple timesteps, we utilized predicted image frames to generate subsequent predictions.


\section{Results on RoAM and discussion }\label{sec:results}
ACPNet is trained on RoAM using the ADAM optimizer \cite{adam}, with a learning rate of 0.0001, a batch size of 8, and $\beta_1=0.5$, $\beta=0.0001$, and $\alpha=1.0$. The dataset comprised 25 video and control-actions sequences of varying lengths, recorded on different days and locations. We divided the dataset into training and test sets, with a ratio of 20:5, respectively. During training and testing, each of the 25 video sequences was segmented into smaller clips of 50 frames. We also maintained a gap of 10 frames between each clip during both training and testing time so that each clip is different from one another. The network was trained to predict 10 future frames based on the past 5 image frames of size $64\times64\times3$ and the history of control actions. However, during inference, we generate 20 future frames while conditioning on the last 5 known image frames and their corresponding action sequences. Moreover, to evaluate ACPNet's performance compared to other state-of-the-art architectures for partially observable scenarios, we trained VANet and MCNet using ADAM with similar hyper-parameters as described in \cite{sarkar2021} for their training on KITTI dataset. Each network is trained for 150,000 iterations on a GTX 3090 GPU-enabled server.

To conduct a quantitative analysis of the performance, we employed four evaluation metrics: Peak Signal-to-Noise Ratio (PSNR), Structural Similarity (SSIM), VGG16 Cosine Similarity \cite{VGG16}, and Fréchet Video Distance (FVD) \cite{FVD}. Among these metrics, FVD measures the spatio-temporal perturbations of the generated videos as a whole, with respect to the ground truth, based on the Fréchet Inception Distance (FID) that is commonly used for evaluating the quality of images from generative frameworks. For frame-wise evaluation, we provided comparative performance plots for VGG16 cosine similarity index, SSIM, and PSNR. The VGG16 cosine similarity index measures the cosine similarity between flattened high-level feature vectors extracted from the VGG network \cite{VGG16}, providing insights into the perceptual-level differences between the generated and ground truth video frames. PSNR and SSIM are widely used frame-level similarity indexes in the current literature.

\begin{enumerate} 
\item \textbf{Quantitative Evaluation:} The figures presenting the quantitative performance metrics for VGG16 similarity, Fréchet Video Distance (FVD) score, Structural Similarity Index (SSIM), and Peak Signal-to-Noise Ratio (PSNR) are displayed in Figure \ref{fig:roam_quant}. Notably, a lower FVD score indicates superior performance, whereas higher values for the remaining three metrics are indicative of better performance. The superiority of ACPNet over VANet and MCNet is evident from the performance plots of VGG16 cosine similarity, as presented in Figure \ref{fig:roam_quant_vgg16}. This observation is also apparent from the performance plots of SSIM similarity and PSNR index, as depicted in Figure \ref{fig:roam_quant_ssim} and Figure \ref{fig:roam_quant_psnr}, respectively. Figure \ref{fig:roam_quant_fvd} reveals that VANet outperforms ACPNet by a marginal amount only in the case of the FVD score. This difference could be attributed to the comparatively lower resolution of the image sequences used for the analysis. Specifically, our experiments involved training and testing the networks on image sequences of size $64\times 64\times 3$. Conducting training and inference with higher-resolution images might offer improved insights into the performance differences observed. In the case of the FVD scores both, all the 3 networks generate a median score below 600. 
\begin{figure*}[ht]
\resizebox{\textwidth}{!}{\begin{tabular}{cccccccccccc}
 & t=1 & t=3 & t=5 & t=6 & t=7 & t=8 & t=13 & t=14 & t=15 & t=20\\
Ground Truth & \imgcell{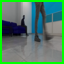} & \imgcell{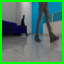} & \imgcell{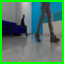}&  \imgcell{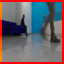}&\imgcell{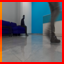} & \imgcell{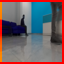} & \imgcell{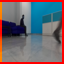} & \imgcell{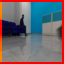} & \imgcell{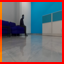} & \imgcell{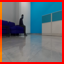} \\
&&&ACPNet&  \imgcell{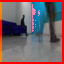}&\imgcell{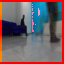} & \imgcell{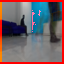} & \imgcell{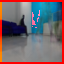} & \imgcell{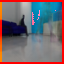} & \imgcell{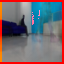} & \imgcell{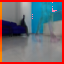} \\
&&&VANet&  \imgcell{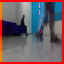}&\imgcell{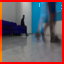} & \imgcell{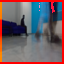} & \imgcell{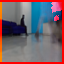} & \imgcell{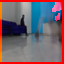} & \imgcell{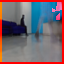} & \imgcell{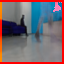} \\
&&&MCNet&  \imgcell{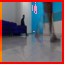}&\imgcell{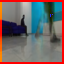} & \imgcell{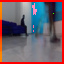} & \imgcell{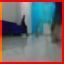} & \imgcell{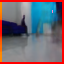} & \imgcell{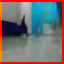} & \imgcell{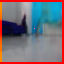}
\end{tabular}}
\caption{Predictions on partially observable RoAM dataset by ACPNet, VANet and MCNet. The models were trained using 5 input-action frames to predict the next ten frames. During the inference phase, the predicted frames were increased to 20}
\label{fig:roam_qual}
\end{figure*}
The plots presented in Figure \ref{fig:roam_quant_vgg16}, Figure \ref{fig:roam_quant_ssim}, and Figure \ref{fig:roam_quant_psnr} for VGG16 cosine similarity, SSIM, and PSNR, respectively, also reveal that MCNet faces considerable difficulties in generating stable predictions for complex scenarios in the RoAM dataset. This is evident from the erratic saw-tooth-like patterns observed in the MCNet plots during inference. 

\item \textbf{Qualitative Evaluation:} Figure \ref{fig:roam_qual} contains examples of raw video frames from the test set, generated by ACPNet, VANet, and MCNet. Upon close observation of the image frames presented in Figure \ref{fig:roam_qual}, it becomes apparent that the predicted frames from VANet and MCNet exhibit significantly more defects compared to those generated by ACPNet. Specifically, MCNet fails to approximate and predict the motion of the leg properly, as evidenced by the frames at timestep $t=13, 14$. Similarly, VANet generates predictions with large pink patches visible in frames at timestep $t=13, 14, 15$, and so on. Although ACPNet generates predictions with some pink patches, these artifacts are much smaller than those from VANet, and ACPNet approximates human motion more accurately compared to the other two frameworks.

\item  \textbf{Discussion}
Based on both quantitative and qualitative analyses presented above, it is evident that ACPNet outperforms VANet and MCNet in making predictions with the new RoAM dataset. Notably, ACPNet has a smaller parametric size than VANet since it does not utilize the acceleration or second-order motion flow encoder. The superior performance of ACPNet, even with a smaller parametric size, can be attributed to the incorporation of the control action vector of the robot in the augmented current image and motion flow maps during training in (\ref{eq:x_hat}) and (\ref{eq:o_hat}). This action conditioning in the training step of (\ref{eq:ACPNet_mod2}), helps the network effectively learn and detangle the effects of robot dynamics on predicted image frames. Additionally, while ACPNet has a similar number of parameters as MCNet, MCNet was not able to learn the complex and partially observable dynamics in the RoAM data even after training for 150K iterations. Thus, these results in Figure \ref{fig:roam_quant} and Figure \ref{fig:roam_qual} further underscore the importance of action-conditioned datasets like RoAM for vision-based prediction and planning frameworks.

Although the proposed prediction framework, ACPNet, is based on a deterministic architecture, the dataset can readily be adapted and trained using stochastic or variational architectures such as Variational Autoencoders (VAEs). The action-conditioned BAIR push dataset has already been utilized and evaluated by various variational architectures, as reported in \cite{finn2} and \cite{LeeICLR2018}. Similar variational frameworks for partially observable scenarios can be designed and adopted for training on RoAM dataset. 

Furthermore, we would like to emphasize that although we collected the dataset using a Turtlebot3 robot due to its  easy accessibility in the robotics community, it can be utilized by any non-holonomic mobile robot of a similar size that employs a 2-wheeled differential drive. This is because the dataset contains control actions in the form of forward velocity and turn rate, which represent the simplest control commands for any mobile robot. For 4-wheel differential drive systems, the RoAM dataset can still be utilized in the pretraining phase with some additional training and fine-tuning in the later stages. We are currently in the process of open-sourcing the entire RoAM dataset to the robotics and computer vision community. We aim to provide access to this valuable dataset to researchers and practitioners in these fields, which will facilitate the development and evaluation of new algorithms and approaches for robot perception, control, and navigation. The partial samples from the dataset can be accessed here: \href{https://tinyurl.com/RoAMData}{https://tinyurl.com/RoAMData}
\end{enumerate}
\section{CONCLUSIONS}\label{sec:conclusion}
In summary, we have introduced the RoAM dataset, which is a novel action-conditioned indoor robot motion dataset containing synchronized control action data, stereo image data, depth disparity maps, image timestamp data, IMU, and odometry data. We have also presented a novel action-conditioned visual prediction framework, ACPNet, and compared its performance on RoAM with two other prediction networks, VANet and MCNet. The superior performance of ACPNet on RoAM highlights the importance of such datasets for investigating the correlation between the dynamics of a mobile agent and the observed image frames by the robot, leading to the development of better prediction and planning frameworks for autonomous navigation.

Currently, the RoAM dataset mainly consists of human walking, sitting, and standing motions. However, in the future, we plan to expand the dataset to include more generalized scenarios and natural human actions, such as bending to pick up an object from the floor, sitting down to examine something, jumping, making a sudden turn, or abruptly stopping. This expansion will enable the development and evaluation of algorithms for more complex and diverse scenarios between humans and robots.

\section*{ACKNOWLEDGMENT}
This research was funded by the Intel India PhD Research Fellowship 2021-2022. We would like to thank Onkar and Archana for their continuous assistance in the data collection process and the other members of GCDSL for their support. The authors would also like to express their appreciation to the Chairman, the administrative staff members, other faculty members, and students of the Department of Aerospace Engineering for their cooperation in the data collection exercise. 

\bibliography{IEEEexample}
\bibliographystyle{IEEEtran}
\addtolength{\textheight}{-12cm}   





\end{document}

%% file: math_command.tex
\usepackage{amsmath,amsfonts,bm}

\def\eqref#1{equation~\ref{#1}}

\def\1{\bm{1}}

\DeclareMathAlphabet{\mathsfit}{\encodingdefault}{\sfdefault}{m}{sl}
\SetMathAlphabet{\mathsfit}{bold}{\encodingdefault}{\sfdefault}{bx}{n}

\def\sF{{\mathbb{F}}}

\def\sR{{\mathbb{R}}}

\newcommand{\image}{\text{image}}

%% file: root.bbl
\begin{thebibliography}{10}
\providecommand{\url}[1]{#1}
\csname url@rmstyle\endcsname
\providecommand{\newblock}{\relax}
\providecommand{\bibinfo}[2]{#2}
\providecommand\BIBentrySTDinterwordspacing{\spaceskip=0pt\relax}
\providecommand\BIBentryALTinterwordstretchfactor{4}
\providecommand\BIBentryALTinterwordspacing{\spaceskip=\fontdimen2\font plus
\BIBentryALTinterwordstretchfactor\fontdimen3\font minus
  \fontdimen4\font\relax}
\providecommand\BIBforeignlanguage[2]{{%
\expandafter\ifx\csname l@#1\endcsname\relax
\typeout{** WARNING: IEEEtran.bst: No hyphenation pattern has been}%
\typeout{** loaded for the language `#1'. Using the pattern for}%
\typeout{** the default language instead.}%
\else
\language=\csname l@#1\endcsname
\fi
#2}}

\bibitem{dataset_imbalance1_2020}
\BIBentryALTinterwordspacing
A.~J. Larrazabal, N.~Nieto, V.~Peterson, D.~H. Milone, and E.~Ferrante,
  ``Gender imbalance in medical imaging datasets produces biased classifiers
  for computer-aided diagnosis,'' \emph{Proceedings of the National Academy of
  Sciences}, vol. 117, no.~23, pp. 12\,592--12\,594, 2020. [Online]. Available:
  \url{https://www.pnas.org/doi/abs/10.1073/pnas.1919012117}
\BIBentrySTDinterwordspacing

\bibitem{dataset_balance2_2017}
J.~Wang, Y.~Chen, S.~Hao, W.~Feng, and Z.~Shen, ``Balanced distribution
  adaptation for transfer learning,'' in \emph{2017 IEEE International
  Conference on Data Mining (ICDM)}, 2017, pp. 1129--1134.

\bibitem{Vaswani_attention2017}
A.~Vaswani, N.~Shazeer, N.~Parmar, J.~Uszkoreit, L.~Jones, A.~N. Gomez, L.~u.
  Kaiser, and I.~Polosukhin, ``Attention is all you need,'' in \emph{Advances
  in Neural Information Processing Systems}, I.~Guyon, U.~V. Luxburg,
  S.~Bengio, H.~Wallach, R.~Fergus, S.~Vishwanathan, and R.~Garnett, Eds.,
  vol.~30.\hskip 1em plus 0.5em minus 0.4em\relax Curran Associates, Inc.,
  2017.

\bibitem{villegasNeurIPS2019}
R.~Villegas, A.~Pathak, H.~Kannan, D.~Erhan, Q.~V. Le, and H.~Lee, ``High
  fidelity video prediction with large stochastic recurrent neural networks,''
  in \emph{In Proceedings of the Thirty-second Advances in Neural Information
  Processing Systems}, ser. NeurIPS 2019.\hskip 1em plus 0.5em minus
  0.4em\relax Curran Associates, Inc., 2019, pp. 81--91.

\bibitem{LiangICCV2017}
X.~Liang, L.~Lee, W.~Dai, and E.~P. Xing, ``Dual motion gan for future-flow
  embedded video prediction,'' in \emph{Proceedings of IEEE International
  Conference on Computer Vision}, ser. ICCV 2017, 2017, pp. 1762--1770.

\bibitem{Denton}
\BIBentryALTinterwordspacing
E.~Denton and R.~Fergus, ``Stochastic video generation with a learned prior,''
  in \emph{Proceedings of the Thirty-fifth International Conference on Machine
  Learning, {ICML 2018}}, ser. Proceedings of Machine Learning Research,
  vol.~80.\hskip 1em plus 0.5em minus 0.4em\relax Stockholmsmässan, Stockholm
  Sweden: PMLR, 10--15 Jul 2018, pp. 1174--1183. [Online]. Available:
  \url{http://proceedings.mlr.press/v80/denton18a.html}
\BIBentrySTDinterwordspacing

\bibitem{BabaeizadehICLR2018}
\BIBentryALTinterwordspacing
M.~Babaeizadeh, C.~Finn, D.~Erhan, R.~H. Campbell, and S.~Levine, ``Stochastic
  variational video prediction,'' in \emph{Proceedings of the Sixth
  International Conference on Learning Representations}, ser. ICLR 2018, 2018.
  [Online]. Available: \url{https://openreview.net/forum?id=rk49Mg-CW}
\BIBentrySTDinterwordspacing

\bibitem{LeeICLR2018}
A.~X. Lee, R.~Zhang, F.~Ebert, P.~Abbeel, C.~Finn, and S.~Levine, ``Stochastic
  adversarial video prediction,'' \emph{arXiv preprint arXiv:1804.01523}, 2018.

\bibitem{CastrejonICCV2019}
L.~Castrejon, N.~Ballas, and A.~Courville, ``Improved conditional vrnns for
  video prediction,'' in \emph{Proceedings of the IEEE/CVF International
  Conference on Computer Vision}, ser. ICCV 2019, October 2019.

\bibitem{GaoCVPR2020}
H.~Gao, H.~Xu, Q.-Z. Cai, R.~Wang, F.~Yu, and T.~Darrell, ``Disentangling
  propagation and generation for video prediction,'' in \emph{In Proc.. of the
  IEEE/CVF International Conference on Computer Vision}, ser. ICCV 2019,
  October 2019.

\bibitem{slrvp2020}
J.-Y. Franceschi, E.~Delasalles, M.~Chen, S.~Lamprier, and P.~Gallinari,
  ``Stochastic latent residual video prediction,'' in \emph{International
  Conference on Machine Learning}.\hskip 1em plus 0.5em minus 0.4em\relax PMLR,
  2020, pp. 3233--3246.

\bibitem{video_transformer2022}
X.~Ye and G.-A. Bilodeau, ``Vptr: Efficient transformers for video
  prediction,'' in \emph{2022 26th International Conference on Pattern
  Recognition (ICPR)}, 2022, pp. 3492--3499.

\bibitem{sarkar2021}
M.~Sarkar, D.~Ghose, and A.~Bala, ``Decomposing camera and object motion for an
  improved video sequence prediction,'' in \emph{NeurIPS 2020 Workshop on
  Pre-registration in Machine Learning}.\hskip 1em plus 0.5em minus 0.4em\relax
  PMLR, 2021, pp. 358--374.

\bibitem{villegas}
\BIBentryALTinterwordspacing
R.~Villegas, J.~Yang, S.~Hong, X.~Lin, and H.~Lee, ``Decomposing motion and
  content for natural video sequence prediction,'' in \emph{Proceedings of the
  Fifth International Conference on Learning Representations}, ser. ICLR-2017,
  Toulon, France, 2017. [Online]. Available:
  \url{http://arxiv.org/abs/1706.08033}
\BIBentrySTDinterwordspacing

\bibitem{KITTI}
A.~Geiger, P.~Lenz, C.~Stiller, and R.~Urtasun, ``Vision meets robotics: The
  kitti dataset,'' \emph{The International Journal of Robotics Research},
  vol.~32, no.~11, pp. 1231--1237, 2013.

\bibitem{KITTI-360}
Y.~Liao, J.~Xie, and A.~Geiger, ``{KITTI}-360: A novel dataset and benchmarks
  for urban scene understanding in 2d and 3d,'' \emph{arXiv preprint
  arXiv:2109.13410}, 2021.

\bibitem{A2D2}
J.~Geyer, Y.~Kassahun, M.~Mahmudi, X.~Ricou, R.~Durgesh, A.~S. Chung,
  L.~Hauswald, V.~H. Pham, M.~M{\"u}hlegg, S.~Dorn, T.~Fernandez,
  M.~J{\"a}nicke, S.~Mirashi, C.~Savani, M.~Sturm, O.~Vorobiov, M.~Oelker,
  S.~Garreis, and P.~Schuberth, ``{A2D2: Audi Autonomous Driving Dataset},''
  2020.

\bibitem{srivastava}
N.~Srivastava, E.~Mansimov, and R.~Salakhudinov, ``Unsupervised learning of
  video representations using lstms,'' in \emph{Proceedings of Thirty-second
  International Conference on Machine Learning}, ser. ICML 2015, Lille, France,
  2015, pp. 843--852.

\bibitem{UCF-101}
K.~Soomro, A.~R. Zamir, and M.~Shah, ``Ucf101: A dataset of 101 human actions
  classes from videos in the wild,'' \emph{arXiv preprint arXiv:1212.0402},
  2012.

\bibitem{HMDB-51}
H.~Kuehne, H.~Jhuang, E.~Garrote, T.~Poggio, and T.~Serre, ``Hmdb: a large
  video database for human motion recognition,'' in \emph{2011 International
  conference on computer vision}.\hskip 1em plus 0.5em minus 0.4em\relax IEEE,
  2011, pp. 2556--2563.

\bibitem{xing}
S.~H.~I. Xingjian, Z.~C.~H. Wang, D.-Y. Yeung, W.-K. Wong, and W.~chun Woo,
  ``Convolutional lstm network: A machine learning approach for precipitation
  nowcasting,'' in \emph{Proceedings of Twenty-nineth Conference on Neural
  Information Processing Systems}, ser. NIPS 2015, Montréal Canada, 2015, pp.
  802--810.

\bibitem{kth}
C.~{Schuldt}, I.~{Laptev}, and B.~{Caputo}, ``Recognizing human actions: a
  local svm approach,'' in \emph{Proceedings of the 17th International
  Conference on Pattern Recognition}, ser. ICPR 2004, vol.~3, 2004, pp. 32--36.

\bibitem{finn2}
C.~Finn and S.~Levine, ``Deep visual foresight for planning robot motion,'' in
  \emph{Proceedings of IEEE International Conference on Robotics and
  Automation}, ser. ICRA 2017, Singapore, May 2017, pp. 2786--2793.

\bibitem{CalTech-pedestrian}
P.~Dollar, C.~Wojek, B.~Schiele, and P.~Perona, ``Pedestrian detection: An
  evaluation of the state of the art,'' \emph{IEEE transactions on pattern
  analysis and machine intelligence}, vol.~34, no.~4, pp. 743--761, 2011.

\bibitem{Stanford-Go}
N.~Hirose, A.~Sadeghian, M.~V{\'a}zquez, P.~Goebel, and S.~Savarese, ``Gonet: A
  semi-supervised deep learning approach for traversability estimation,'' pp.
  3044--3051, 2018.

\bibitem{ROS}
\BIBentryALTinterwordspacing
{Stanford Artificial Intelligence Laboratory et al.}, ``Robotic operating
  system.'' [Online]. Available: \url{https://www.ros.org}
\BIBentrySTDinterwordspacing

\bibitem{sarkar-cis2023}
M.~Sarkar, M.~Prabhakar, and D.~Ghose, ``Avoiding obstacles with geometric
  constraints on lidar data for autonomous robots,'' in \emph{Third Congress
  on Intelligent Systems}, S.~Kumar, H.~Sharma, K.~Balachandran, J.~H. Kim, and
  J.~C. Bansal, Eds.\hskip 1em plus 0.5em minus 0.4em\relax Singapore: Springer
  Nature Singapore, 2023, pp. 749--761.

\bibitem{michels_2005}
\BIBentryALTinterwordspacing
J.~Michels, A.~Saxena, and A.~Y. Ng, ``High speed obstacle avoidance using
  monocular vision and reinforcement learning,'' in \emph{Proceedings of the
  22nd International Conference on Machine Learning}, ser. ICML '05.\hskip 1em
  plus 0.5em minus 0.4em\relax New York, NY, USA: Association for Computing
  Machinery, 2005, p. 593–600. [Online]. Available:
  \url{https://doi.org/10.1145/1102351.1102426}
\BIBentrySTDinterwordspacing

\bibitem{collision_cone}
A.~Chakravarthy and D.~Ghose, ``Obstacle avoidance in a dynamic environment: a
  collision cone approach,'' \emph{IEEE Transactions on Systems, Man, and
  Cybernetics - Part A: Systems and Humans}, vol.~28, no.~5, pp. 562--574,
  1998.

\bibitem{convlstm}
X.~Shi, Z.~Chen, H.~Wang, D.-Y. Yeung, W.~kin Wong, and W.~chun WOO,
  ``Convolutional {LSTM} network: A machine learning approach for precipitation
  nowcasting,'' in \emph{Proceedings of the Twenty-ninth International
  Conference on Neural Information Processing Systems}, ser. NIPS 2015,
  Montreal, 2015, pp. 802--810.

\bibitem{adam}
D.~P. Kingma and J.~Ba, ``Adam: A method for stochastic optimization,''
  \emph{CoRR}, vol. abs/1412.6980, 2015.

\bibitem{VGG16}
K.~Simonyan and A.~Zisserman, ``Very deep convolutional networks for
  large-scale image recognition,'' in \emph{3rd International Conference on
  Learning Representations, {ICLR} 2015, San Diego, CA, USA, May 7-9, 2015,
  Conference Track Proceedings}, Y.~Bengio and Y.~LeCun, Eds., 2015.

\bibitem{FVD}
T.~Unterthiner, S.~van Steenkiste, K.~Kurach, R.~Marinier, M.~Michalski, and
  S.~Gelly, ``Towards accurate generative models of video: A new metric \&
  challenges,'' \emph{arXiv preprint arXiv:1812.01717}, 2018.

\end{thebibliography}
